\documentclass{article}

\usepackage{arxiv}

\usepackage[utf8]{inputenc} 
\usepackage[T1]{fontenc}    
\usepackage{hyperref}       
\usepackage{url}            
\usepackage{booktabs}       
\usepackage{amsfonts}       
\usepackage{nicefrac}       
\usepackage{microtype}      
\usepackage{graphicx}
\usepackage[numbers]{natbib}
\usepackage{doi}
\usepackage{amsmath}
\usepackage{hyperref}
\usepackage{multirow}
\usepackage{lipsum}
\usepackage{tablefootnote}
\usepackage{makecell}
\usepackage{mathtools}
\usepackage{graphicx}
\usepackage{caption}

\title{Kolmogorov-Arnold Convolutions: Design Principles and Empirical Studies}

\author{ \href{https://orcid.org/0000-0003-2634-7857}{\includegraphics[scale=0.06]{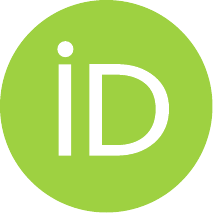}\hspace{1mm}Ivan Drokin} \\
	Deep Learning Researcher\\
        Seath the Scaleless Research Group \\
	\texttt{ivan.s.drokin@gmail.com} \\
}

\renewcommand{\shorttitle}{Kolmogorov-Arnold Convolutions: Design Principles and Empirical Studies}

\hypersetup{
pdftitle={A template for the arxiv style},
pdfsubject={cs.LG},
pdfauthor={Ivan Drokin},
pdfkeywords={Convolutional Kolmogorov-Arnold Networks, Kolmogorov-Arnold Networks, Computer Vision},
}

\begin{document}
\maketitle

\begin{abstract}
	The emergence of Kolmogorov-Arnold Networks (KANs) has sparked significant interest and debate within the scientific community. This paper explores the application of KANs in the domain of computer vision (CV). We examine the convolutional version of KANs, considering various nonlinearity options beyond splines, such as Wavelet transforms and a range of polynomials. We propose a parameter-efficient design for Kolmogorov-Arnold convolutional layers and a parameter-efficient finetuning algorithm for pre-trained KAN models, as well as KAN convolutional versions of self-attention and focal modulation layers. We provide empirical evaluations conducted on MNIST, CIFAR10, CIFAR100, Tiny ImageNet, ImageNet1k, and HAM10000 datasets for image classification tasks. Additionally, we explore segmentation tasks, proposing U-Net-like architectures with KAN convolutions, and achieving state-of-the-art results on BUSI, GlaS, and CVC datasets. We summarized all of out finding in a preliminary design guide of KAN convolutional models for computer vision tasks. Furthermore, we investigate regularization techniques for KANs. All experimental code and implementations of convolutional layers and models, pre-trained on ImageNet1k weights are available on GitHub: \href{https://github.com/IvanDrokin/torch-conv-kan}{https://github.com/IvanDrokin/torch-conv-kan}.
\end{abstract}


\section{Introduction}
The rapid evolution of deep learning architectures has significantly advanced the field of computer vision, particularly in tasks that require the analysis of complex spatial data. Convolutional Neural Networks (CNNs), initially proposed by LeCun et al. \cite{5537907}, have become a cornerstone in this domain due to their ability to efficiently process high-dimensional data arrays such as images. These networks typically employ linear transformations followed by activation functions in their convolutional layers to discern spatial relationships, thereby reducing the number of parameters needed to capture intricate patterns in visual data. Since 2012, following the success of AlexNet \cite{NIPS2012_c399862d} in the ImageNet classification challenge, CNNs have dominated the field of computer vision until the emergence of Vision Transformers \cite{dosovitskiy2021an}. Innovations such as Residual Networks \cite{10.1145/2983402.2983406} and Densely Connected networks \cite{8099726}, along with numerous subsequent works, have significantly advanced the achievable quality of models based on convolutional layers, enabling the effective training of very large and deep networks.

In segmentation tasks, especially within the biomedical domain, CNNs have also become foundational with the advent of the U-Net \cite{ronneberger2015unet} architecture, which has subsequently inspired a whole family of U-Net-like architectures for segmentation tasks.

Recent developments in deep learning have seen the integration of sophisticated mathematical theories into neural network architectures, enhancing their capability to handle complex data structures. One such innovation is the Kolmogorov-Arnold Network (KAN) \cite{liu2024kan}, which leverages the Kolmogorov-Arnold theorem to incorporate splines into its architecture, offering a compelling alternative to traditional Multi-Layer Perceptrons (MLPs). Quickly following the original work introducing KANs, modifications, and improvements have emerged that attempt to overcome various issues associated with the spline-based approach, namely computational overhead and a large number of trainable parameters.

In light of these advancements, this paper explores the adaptation of KANs to convolutional layers—a prevalent component in many CNN architectures used for image classification. Traditional CNNs, while effective, often rely on fixed activation functions and linear transformations, which can be enhanced through the flexibility and reduced parametric complexity offered by KANs. In this work, we explore various modifications of the original KAN model, propose efficient designs for convolutional KAN models, conduct empirical evaluations of the proposed approaches, and compare them with classical convolutional networks.

Our contributions can be summarized as follows:

\begin{enumerate}

\item We present the Bottleneck Convolutional Kolmogorov-Arnold layer, which retains the properties of Kolmogorov-Arnold layers while significantly reducing memory requirements.

\item We empirically investigate several modifications of convolutional KAN models in the context of image classification tasks, including studies on regularization and hyperparameter tuning, across multiple datasets.

\item We introduce a parameter-efficient finetuning algorithm for Gram polynomials in Convolutional Kolmogorov-Arnold Networks, which substantially reduces the number of trainable parameters needed when adapting a pre-trained model to new tasks.

\item We redesign U-Net-like models for segmentation tasks with Kolmogorov Arnold Convolutional layers instead if regular convolutions, demonstrating that such models achieve state-of-the-art results on three diverse biomedical datasets.

\item We propose redesigns of Self-Attention and Focal Modulation layers based on Bottleneck Convolutional Kolmogorov-Arnold layers, which significantly improve the performance of classification models.

\item Based on the results of our empirical studies, we formulate design principles for constructing successful computer vision models based on Bottleneck Convolutional Kolmogorov-Arnold layers.

\item We provide the entire codebase on GitHub, and pre-trained weights on ImageNet are available on HuggingFace to accelerate research and ensure reproducibility.

\end{enumerate}

Our work is structured as follows. In Section 2, we present an overview of relevant works on Kolmogorov-Arnold networks. In Section 3, we describe the Kolmogorov-Arnold convolutional layer, its bottleneck version, attention and focal modulation layers, regularization techniques for the Kolmogorov-Arnold convolutional layer, and introduce and describe the PEFT algorithm for polynomial variants of Kolmogorov-Arnold convolutional networks. In Section 4, we present experiments on image classification and segmentation tasks. In Section 5, we conduct an ablation study for the bottleneck Kolmogorov-Arnold convolutional layer. In Section 6, we summarize all experimental results into a design guide for constructing Kolmogorov-Arnold convolutional networks.

\section{Related works}

The application of the Kolmogorov-Arnold theorem in neural networks marks a significant theoretical integration that enhances the expressiveness and efficiency of neural models. The theorem, which provides a way to represent any multivariate continuous function as a composition of univariate functions and additions, has been adapted in the design of Kolmogorov-Arnold Networks (KANs). KANs differ from traditional Multi-Layer Perceptrons (MLPs) by replacing linear weight matrices with learnable splines, thus reducing the number of parameters required and potentially improving the generalization capabilities of the network.

Recent research has proposed several variations of KANs to address specific limitations and enhance their performance.

Fast KAN by Li et al. (2024) introduced an adaptation where B-splines are replaced by Radial Basis Functions (RBFs). This modification aims to reduce the computational overhead associated with splines. The work demonstrated that third-order B-splines used in traditional KANs could be effectively approximated by Gaussian radial basis functions, resulting in FastKAN—a faster implementation of KAN which also functions as an RBF network.

Wavelet-based KANs (Wav-KAN), as presented by Bozorgasl et al. (2024), incorporate wavelet functions into the KAN structure to enhance both interpretability and performance. Wav-KAN leverages the properties of wavelets to efficiently capture high-frequency and low-frequency components of input data, balancing accurate representation of data structures with robustness against overfitting. The implementation employs discrete wavelet transforms (DWT) for multiresolution analysis, which simplifies the computation process. Wav-KAN has demonstrated enhanced accuracy, faster training speeds, and increased robustness compared to traditional Spl-KAN and MLPs.

ReLU-KAN by Qiu et al. (2024) addresses the computational complexity of basis function calculations in traditional KANs by introducing a novel implementation that utilizes ReLU (Rectified Linear Unit) and point-wise multiplication. This approach optimizes the computation process for efficient CUDA computing, achieving a significant speedup (20x) over traditional KANs while maintaining stable training and superior fitting ability. ReLU-KAN preserves the "catastrophic forgetting avoidance" property of KANs, making it a practical choice for both inference and training within existing deep learning frameworks.

Polynomial-based variations of KANs have also been explored. Chebyshev KAN replaces B-splines with Chebyshev polynomials, which are known for their excellent function approximation capabilities and can be calculated recursively. This approach aims to improve the performance and intuitiveness of KANs. Meanwhile, Gram KAN leverages the simplicity of Gram polynomial transformations, characterized by their discrete nature. This discrete approach is particularly suited for handling discretized datasets like images and text data, offering a novel method for data processing in neural networks.

Kolmogorov Arnold Legendre Network (KAL-Net) \cite{torchkan} represents a novel architecture using Legendre polynomials to surpass traditional polynomial approximations like splines in KANs. KAL-Net utilizes Legendre polynomials up to a specific order for input normalization, capturing nonlinear relationships more efficiently. By leveraging caching mechanisms and recurrence relations, KAL-Net enhances computational efficiency. It employs SiLU (Sigmoid Linear Unit) activation functions and layer normalization to stabilize outputs and improve training stability. KAL-Net has demonstrated remarkable accuracy (97.8\%) on the MNIST dataset and efficiency, with an average forward pass taking only 500 microseconds, highlighting its potential in handling complex image patterns and computational efficiency.

Convolutional Kolmogorov-Arnold Networks (Convolutional KANs) \cite{bodner2024convolutional} were introduced as an alternative to standard Convolutional Neural Networks (CNNs) in a recent study. This approach integrates the non-linear activation functions from KANs into convolutional layers, creating a new layer type that maintains similar accuracy levels to traditional CNNs while using half the number of parameters. This significant reduction in parameters highlights a promising direction for optimizing neural network architectures.

In the domain of medical image segmentation, U-Net \cite{ronneberger2015unet} has become a foundational architecture, utilizing an encoder-decoder structure to effectively capture image features. Variations such as U-Net++ \cite{zhou2018unet++} improve segmentation accuracy through nested structures that fuse multi-scale features. Beyond convolution-based methods, transformer-based models like Vision Transformer \cite{dosovitskiy2021an} and TransUNet \cite{chen2021transunettransformersmakestrong} have shown effectiveness in incorporating global context into medical image segmentation.

$U^2$-Net \cite{QIN2020107404}, designed for salient object detection, employs a two-level nested U-structure to capture contextual information across different scales. This architecture increases depth without significantly increasing computational cost, offering competitive performance on various datasets.

Li et al. \cite{li2024ukanmakesstrongbackbone} investigate and redesign the established U-Net pipeline by integrating Kolmogorov-Arnold Network (KAN) layers into the intermediate tokenized representations, resulting in the U-KAN architecture.

\section{Method}
This section details the methods used in our study, focusing on Kolmogorov-Arnold convolutions and their various adaptations and enhancements. The subsections cover the following topics:

In Section \ref{sec:convkans}, we provide a brief overview of Kolmogorov-Arnold convolutions, as initially presented in \cite{bodner2024convolutional}, including their formalization and various basis function options such as splines, Radial-Basis Functions, Wavelets, and polynomials. We also introduce the use of Gram polynomials for parameter-efficient fine-tuning.

Section \ref{sec:bottleneck_kans} addresses the primary issue with Kolmogorov-Arnold Convolutions—the high number of parameters introduced by the basis functions. We propose a bottleneck version to mitigate this problem, involving a squeezing convolution before and an expanding convolution after applying the basis function. This design includes a mixture of experts for effective implementation.

In Section \ref{sec:selfkantention}, we describe the construction of Self-KAGtention layers by substituting traditional convolutions with Kolmogorov-Arnold convolutional layers. Additionally, we introduce Focal KAGN Modulation, where all convolutional layers in the original focal modulation are replaced with Kolmogorov-Arnold convolutional layers.

Section \ref{sec:convreg} discusses various regularization techniques, including weight and activation penalties, dropout placements, and additive Gaussian noise injection. We explore the impact of these techniques on Kolmogorov-Arnold convolutional networks.

Finally in Section \ref{sec:peftmethod}, we introduce the parameter-efficient finetuning algorithm for polynomial variants of Kolmogorov-Arnold convolutional networks. We outline several options for parameter-efficient fine-tuning, aiming to reduce the number of trainable parameters while adapting pre-trained models to new tasks.

\subsection{Kolmogorov-Arnold Convolutions}
\label{sec:convkans}
Kolmogorov-Arnold convolutions were presented in \cite{bodner2024convolutional}, in this section we briefly cover the formalization of Kolmogorov-Arnold convolutions.

Kolmogorov-Arnold Convolutions could be stated as follows: the kernel consists of a set of univariate non-linear functions. Suppose we have an input image $y, y \in R^{c \times n \times m}$, where $c$ is the number of channels, and $n, m$ are the height and width of an image respectively. Then, KAN-based convolutions with kernel size $k$ could be defined as:
\begin{align*}
x_{ij}=\sum_{d=1}^{c}\sum_{a=0}^{k-1}\sum_{b=0}^{k-1}\varphi_{a,b,d}(y_{d, i+a, j+b}); i = \overline{1, n - k + 1}, j = \overline{1, m - k + 1}
\end{align*}

Each $\varphi$ is a univariate non-linear learnable function with its own trainable parameters. In the original paper\cite{liu2024kan}, the authors propose to use this form of the functions:
\begin{gather*}
\varphi = w_b \cdot b(x) + \Tilde{\varphi}(x),
\\
\Tilde{\varphi}(x) = w_s \cdot Spline(x)
\\
b(x) = SiLU(x) = x/(1 + e^{-x})
\end{gather*}

Similar to KANs, other than splines could be chosen as basis function  $\Tilde{\varphi}(x)$: Radial-Basis Function, Wavelets, polynomials, etc.


Replacing splines with Gram polynomials was proposed in \cite{GRAMKAN}. The Gram polynomials, or the discrete Chebyshev polynomial, $t^N_n(x)$ is a polynomial of degree $n$ in $x$, for $n = 0, 1, 2,\ldots, N -1$, constructed such that two polynomials of unequal degree are orthogonal with respect to the weight function
$w(x) = \sum_{r = 0}^{N-1} \delta(x-r),$ with $\delta(\cdot)$ being the Dirac delta function. That is,
 $\int_{-\infty}^{\infty} t^N_n(x) t^N_m (x) w(x) \, dx = 0 \quad \text{ if } \quad n \ne m .$ In the case of splines by Gram polynomials, KAN Convolutions are defined as follows.

 \begin{gather*}
\varphi = w_b \cdot b(x) + \Tilde{\varphi}(x),
\\
\Tilde{\varphi}(x) = \sum_{i = 0}^{N+1} w_i  \cdot t^i_n(x)
\\
b(x) = SiLU(x) = x/(1 + e^{-x})
\label{gram_conv}
\end{gather*}

This reformulation, on one side, allows an option for parameter-efficient fine-tuning of the pre-trained model, and on the other hand, it reduces the number of trainable parameters.

\subsection{Bottleneck Kolmogorov-Arnold Convolutions}
\label{sec:bottleneck_kans}

\begin{figure}[!ht]
    \centering
    \includegraphics[width=0.75\linewidth]{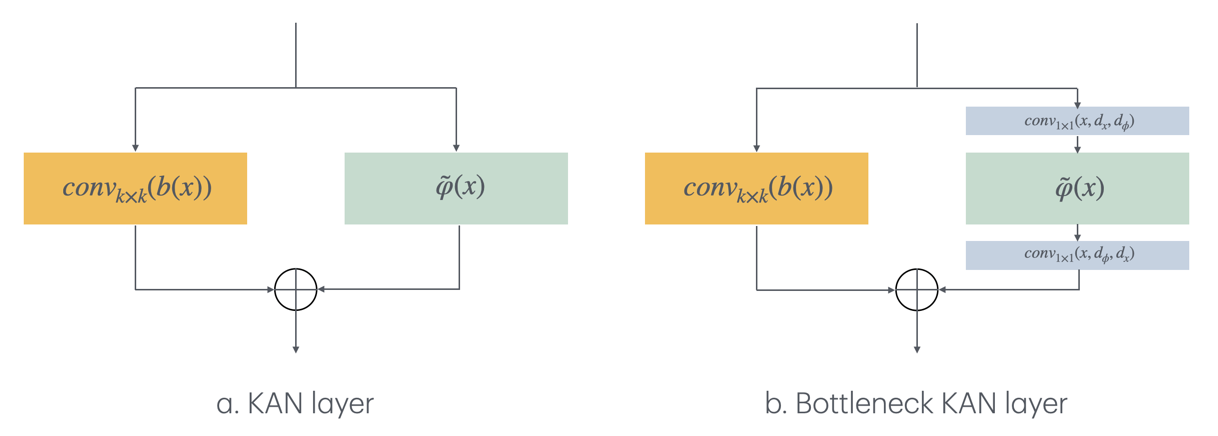}
    \caption{KAN Convolution (left) and Bottleneck KAN Convolution (right). The main difference between these two types of layers is a encoder-decoder convolutional layers on the right data stream.}
    \label{fig:bn_convs}
\end{figure}

The main problem with KAN Convolutions lies in the $Spline$ part of the model. Whatever type of basis function one chooses, the basis introduces a lot of parameters to the model, which leads to higher resource requirements during training and increases the probability of overfitting. To overcome those issues, we propose to use Bottleneck Kolmogorov-Arnold Convolutions (see Fig.\ref{fig:bn_convs}).

Before applying basis to input data, we propose to use squeezing convolution with kernel size equal to 1 before applying basis function to input and expanding convolution with kernel size equal to 1 after. Intuitively, it could be considered a one-layer encoder that helps extract meaningful features from the input before processing it via a chosen basis, and then a one-layer decoder decodes it back. residual activation helps to preserve necessary details that could be lost during encoding and decoding of the input.


\begin{figure}[!ht]
  \begin{minipage}[b]{.45\linewidth}
    \centering
    \includegraphics[width=0.75\linewidth]{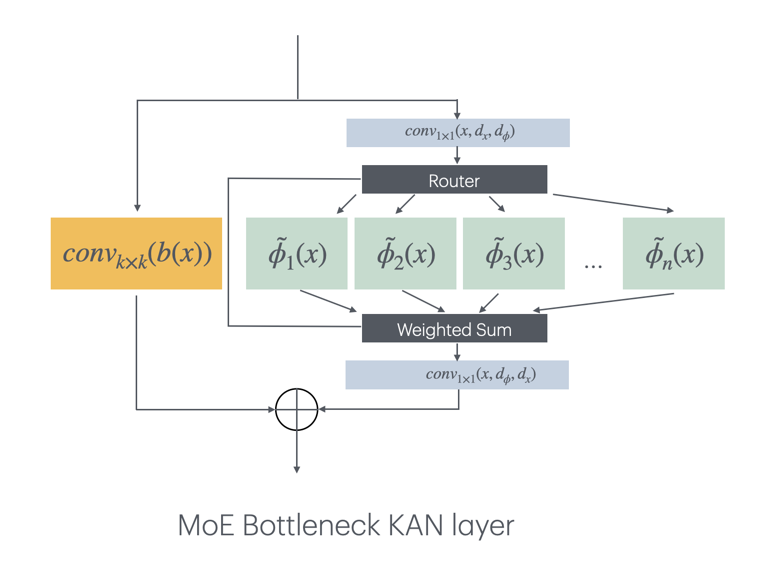}
    \captionof{figure}{Bottleneck Kolmogorov-Arnold Convolutional Mixture of Experts. The router and experts are placed between bottleneck convolutions, and each expert is a $\Tilde{\varphi}$ set of univariate functions. We use sparsely-gated mixture-of-experts \cite{shazeer2017outrageously}.}
    \label{fig:bn_convs_moe}
  \end{minipage}\hfill
  \begin{minipage}[b]{.45\linewidth}
     \centering
    \includegraphics[width=0.75\linewidth]{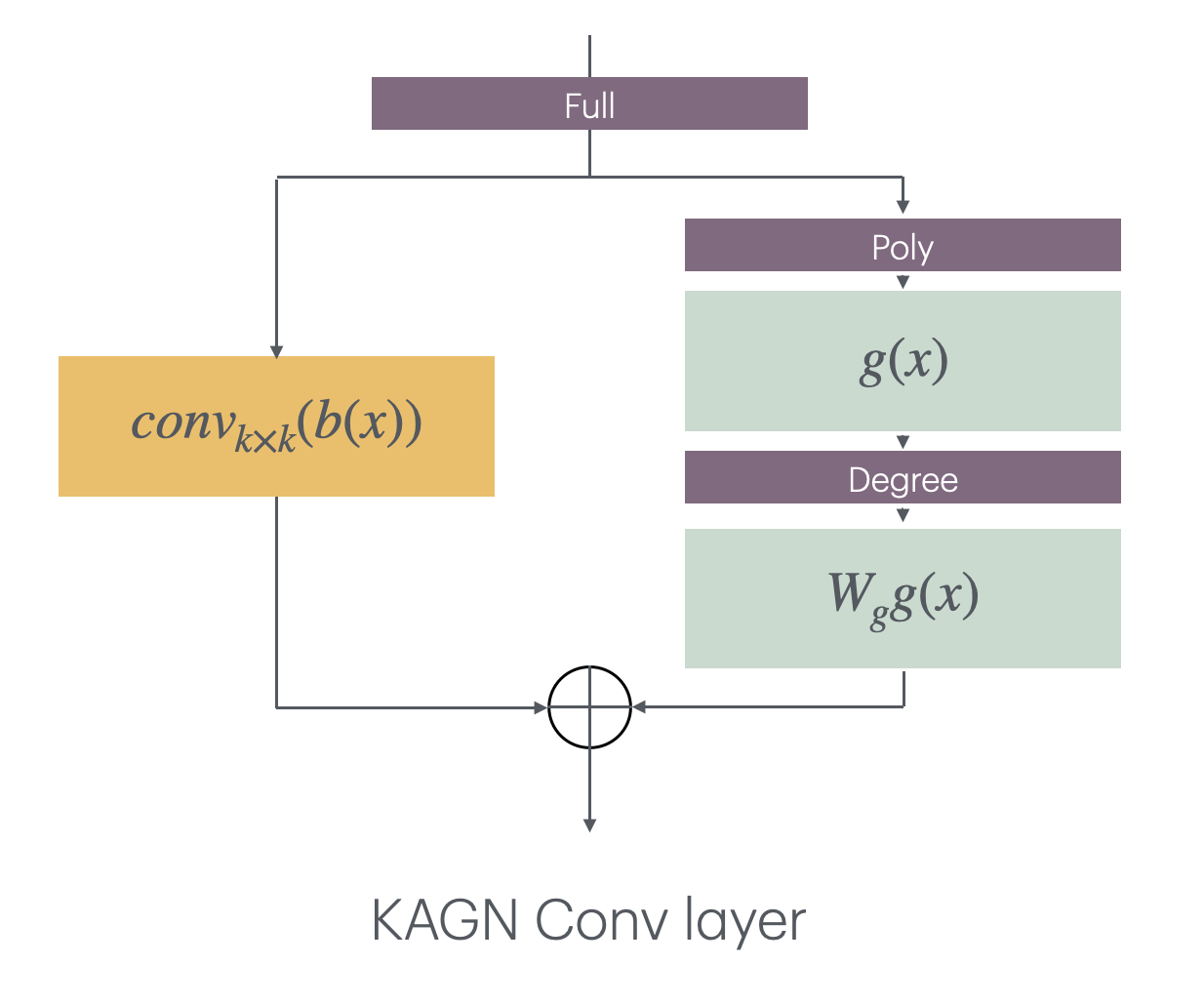}
    \captionof{figure}{Possible dropout layer placement inside KAGN Convolution layer: Full - before the layer, Poly - before computing Gram basis, and Degree - before weighted sum of previously computed basis.}
    \label{fig:conv_kan_drop}
  \end{minipage}
\end{figure}

\begin{figure}[!ht]
    \centering
    \includegraphics[width=0.75\linewidth]{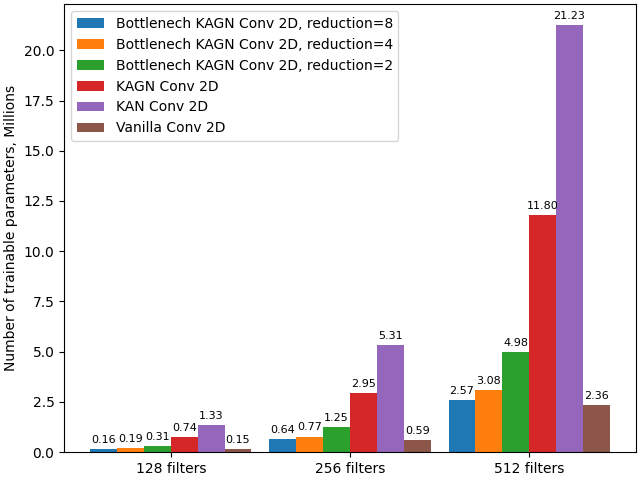}
    \caption{Comparison of the number of trainable parameters for different types of layers. The reduction parameter for Bottleneck KAGN Conv 2D is responsible for the ratio of $d_x$ to $d_\varphi$}
    \label{fig:param_vs_param}
\end{figure}

We compare the number of trainable parameters across various layer types (see Fig.\ref{fig:param_vs_param}). The reduction parameter in the Bottleneck KAGN Conv 2D layer controls the ratio of $d_x$ and $d_\phi$: $reduction = d_x / d_\varphi$. This parameter is crucial for determining the efficiency and scalability of the model. By adjusting the reduction parameter, we can balance the complexity and performance of the model, emphasizing the importance of careful parameter management in deep learning architectures.

Such design allows us to implement a mixture of experts effectively. We utilize sparsely-gated mixture-of-experts \cite{shazeer2017outrageously} in-between the encoder and decoder convolutions, with a set of $\Tilde{\varphi_i}(x)$ as experts (see Fig.\ref{fig:bn_convs_moe}).

\subsection{Kolmogorov-Arnold Sefl-Attention and Focal Modulation}
\label{sec:selfkantention}

In \cite{9710031}, a self-attention layer was introduced where the K, Q, and V projections were replaced by convolutional layers instead of linear projections. In this paper, we propose a similar approach to construct Self-KAGtention layers by substituting traditional convolutions with KAN convolutional layers. Given that the self-attention operation requires $O(n^2)$ memory, where $n$ is the number of tokens or tensor pixels in convolutional models, we suggest using optional bottleneck convolutions with a  $1 \times  1$ kernel and placing the self-attention layer between these two convolutions. We call this version Self-KAGNtention.

Yang et al. \cite{yang2022focal} introduce the Focal Modulation layer to serve as a seamless replacement for the Self-Attention Layer. The layer boasts high interpretability, making it a valuable tool for Deep Learning practitioners. Here we propose a Focal KAGN Modulation, where all convolutional layers from the original focal modulation are replaced with KAN convolutional layers. As the Focal Modulation layer uses grouped convolutions in the hierarchical contextualization stream of data processing with the number of groups equals to the filter number, we note that in this case, Bottleneck KAN Convolutions should be replaced by KAN Convolutions.

\subsection{Regularizations in Kolmogorov-Arnold Convolutional}
\label{sec:convreg}

Applying regularization techniques to Kolmogorov-Arnold Convolutional involves straightforward weight and activation penalties. However, dropout requires careful consideration of its placement. Let's describe the polynomial version of Kolmogorov-Arnold Convolutional. Instead of computing splines over x, we could use Gram, Chebyshev, Legendre, and other Polynomials. In this case, we first need to compute the polynomial basis over x, and then perform the weighted sum of. (Fig.\ref{fig:conv_kan_drop}).
In that case, we have 3 possible options for dropout placement: before the layer (we will refer to this position as "Full"), before polynomial basis calculation (we will refer to this position as "Poly"), and before weight application to polynomials (we will refer to this position as "Degree"). 


The authors of \cite{liu2024kan} state the benefits of KANs over MLPs, and one of them was robustness to noise and adversarial attacks. From this observation, we could derive an alternative way of regularization. Instead of zeroing out some fraction of neurons, we could add additive Gaussian noise to a layer’s input, thus forcing the model to filter this noise and be more robust against noise in unseen data. Similar to dropout, there are three placements available: "Full", "Poly", and "Degree". More formal, for a given neuron $y_i = f(w_i \cdot x_i)$, during noise injection, the neuron output is:
\begin{gather*}
y_i' = \begin{cases}
y_i + \alpha\varepsilon_i, \varepsilon_i\sim N(0, \sigma^2(y)) & \text{with probability } p \\ 
y_i & \text{with probability } 1-p
\end{cases}
\end{gather*}
In this equation, $\alpha$ is a parameter that controls the amount of added noise, and $\sigma^2(y$) is a variance of the input, computed for each input channel.

\subsection{Parameter Efficient Finetuning}
\label{sec:peftmethod}
Let's assume we have a pre-trained model $L$ with Gram KAN convolution layers and we want to fine-tune this model on downstream tasks, e.g. classification dataset $D = \{x_i, y_i\}_{i=1}^{T}$. Let $G_i(w^i_b, w^i_g)$ be a $i$-th layer of the model $L$, and $w^i_b$ are the weights of residual activation and $w^i_g = \{w^i_j\}_{j=0}^{N+1}$ are the weights of Gram polynomials (see formula \ref{gram_conv}). Assume that model $L$ has $M$ Gram KAN convolutional layers.

Then we have several options for parameter-efficient fine-tuning for a downstream task.

\begin{itemize}
    \item Only $\Tilde{w}_g = \{w^i_j\}_{j=R}^{N+1}, R > 1$ are trainable parameters, while all other parameters of $G_i(w^i_b, w^i_g)$ for all $i$ are frozen.
    \item New parameter $\Tilde{w^i}_{N+2}$ could be introduced and the Gram basis will be expanded by one (or more) polynomial of higher degree.
\end{itemize}

Then we can formalize the Parameter Efficient Finetuning algorithm for Gram KAN convolutional models as follows.

\begin{gather*}
\mathcal{C}(L(x_t), y_t) \xrightarrow[\{w^i_j\}_{j=R_l}^{R_u}, i=1 \dots M]{t=1 \dots T} min
\end{gather*}

Where $\mathcal{C}$ is a loss function, $R_l$ and $R_u$ define the minimum and maximum degrees of Gram polynomials, respectively, at which the corresponding weights will be updated. In case when $R_u > N+1$ we add additional polynomials to the basis and initialize them with zeros.

These fine-tuning options sufficiently reduce the number of trainable parameters. Intuitively, one could consider this method as a refining of high-order features to match new data distribution. Empirical evaluation of this method is presented in section \ref{sec:peft}.

\section{Experiments}

In this section, we present the following experiments. \hyperref[sec:baseline]{Section 4.1} includes experiments on the MNIST \cite{deng2012mnist}, CIFAR10, and CIFAR100 \cite{Krizhevsky2009LearningML} datasets with different formalizations of KAN convolutions. \hyperref[sec:reg]{Section 4.2} presents experiments with various regularizations and hyperparameter optimization. \hyperref[sec:cifar100]{Section 4.3} provides results for Bottleneck convolutions on CIFAR100 and Tiny ImageNet \cite{Le2015TinyIV} and also considers ResNet-like and DenseNet-like architectures. \hyperref[sec:selfatt]{Section 4.4} presents results of self-attention layers experiments. \hyperref[sec:in1k]{Section 4.5} presents results for ImageNet1k dataset, and \hyperref[sec:peft]{Section 4.6} presents results of parameter-efficient finetuning. \hyperref[sec:seg]{Section 4.7} presents results for segmentation tasks.

\subsection{Baseline on MNIST, CIFAR10 and CIFAR100}
\label{sec:baseline}

Baseline models were chosen to be simple networks with 4 and 8 convolutional layers. To reduce the feature's spatial dimensions, convolutions with dilation=2 were used. In the 4-layer model, the second and third convolutions had dilation=2, while in the 8-layer model, the second, third, and sixth convolutions had dilation=2.

The number of channels in the convolutions was the same for all models.
\begin{itemize}
    \item For 4 layers: 32, 64, 128, 512
    \item For 8 layers: 2, 64, 128, 512, 1024, 1024, 1024, 1024
\end{itemize}

After the convolutions, Global Average Pooling was applied, followed by a linear output layer.

In the case of classic convolutions, a traditional structure was used: convolution - batch normalization - ReLU.
In the case of KAN convolutions, after KAN convolution layer batch normalization and SiLU are applied. All experiments were conducted on an NVIDIA RTX 3090 with identical training parameters.

In this section, we are investigating performance of several KANs options: spline-based (KANConv), RBF-based (FastKANConv), Legendre polynomials version (KALNConv), Chebyshev polynomials (KACNConv), Gram polynomials (KAGNConv) and Wavelet KANs (WavKANConv).

\begin{table}[!ht]
    \centering
    \resizebox{\textwidth}{!}{\begin{tabular}{l||c|c|c||c|c|c||c|c|c}
    \toprule
        \multirow{2}{*}{\textbf{Model}} & \multicolumn{3}{c||}{\textbf{MNIST}} & \multicolumn{3}{c||}{\textbf{CIFAR10}} & \multicolumn{3}{c}{\textbf{CIFAR100}}\\ \cline {2-10}
        & Val. Accuracy & Params., M & Eval Time, s & Val. Accuracy & Params.,  & Eval Time, s & Val. Accuracy & Params.,  & Eval Time, s \\ \hline
        Conv, 4 layers, baseline & 99.42 & 0.1 & 0.7008 & 73.18 & 0.1 & 1.8321 & 42.29 & 0.12 & 1.5994 \\ \hline
        KANConv, 4 layers & 99.00 & 3.49 & 2.6401 & 52.08 & 3.49 & 3.7972 & 21.78 & 3.52 & 4.0262 \\ \hline
        FastKANConv, 4 layers & 97.65 & 3.49 & 1.5999 & 64.95 & 3.49 & 2.3716 & 34.32 & 3.52 & 2.7457 \\ \hline
        KALNConv, 4 layers & 84.85 & 1.94 & 1.7205 & 10.28 & 1.94 & 3.0527 & 5.97 & 1.97 & 3.0919 \\ \hline
        KACNConv, 4 layers & 97.62 & 3.92 & 1.6710 & 52.01 & 3.92 & 2.3972 & 23.17 & 0.42 & 2.6522 \\ \hline
        KAGNConv, 4 layers & \textbf{99.49} & 0.49 & 1.7253 & 65.84 & 0.49 & 2.2570 & \textbf{47.36} & 1.97 & 2.3399 \\ \hline
        WavKANConv, 4 layers & 99.23 & 0.95 & 7.4622 & \textbf{73.63} & 0.95 & 11.2276 & 41.50 & 0.98 & 11.4744 \\ \midrule
        Conv, 8 layers, baseline & 99.63 & 1.14 & 1.2061 & 83.05 & 1.14 & 1.8258 & 57.52 & 1.19 & 1.8265 \\ \hline
        KANConv, 8 layers & 99.37 & 40.7 & 4.2011 & 74.66 & 40.7 & 5.4858 & 36.18 & 40.74 & 5.7067 \\ \hline
        FastKANConv, 8 layers & 99.49 & 40.7 & 2.1653 & 74.66 & 40.7 & 5.4858 & 43.32 & 40.74 & 2.7771 \\ \hline
        KALNConv, 8 layers & 49.97 & 22.61 & 1.7815 & 15.97 & 22.61 & 2.7348 & 1.74 & 22.65 & 2.6863 \\ \hline
        KACNConv, 8 layers & 99.32 & 18.09 & 1.6973 & 62.14 & 18.09 & 2.3459 & 25.01 & 18.14 & 2.3826 \\ \hline
        KAGNConv, 8 layers & \textbf{99.68} & 22.61 & 2.2402 & 84.14 & 22.61 & 2.5849 & \textbf{59.27} & 22.66 & 2.6460 \\ \hline
        WavKANConv, 8 layers & 99.57 & 10.73 & 59.1734 & \textbf{85.37} & 10.73 & 28.0385 & 55.43 & 10.78 & 30.5438 \\
    \end{tabular}}
    \caption{Results on MNIST, CIFAR10, and CIFAR100 datasets}
    \label{cifar100_baseline}
\end{table}

As we can see from the Table \ref{cifar100_baseline}, Gram polynomials-based and Wavelet-based versions perform better than other other options, and outperform vanilla convolutions. Due to wavelet-based KANs' higher computational resource requirements, we will focus on Gram KANs as the main basis function option in further research.

\subsection{Regularization study and hyperparameters optimization}
\label{sec:reg}
\subsubsection{Regularization study}
Baseline model was chosen to be simple networks with 8 convolutional layers with Gram polynomials as basis functions. To reduce the feature's spatial dimensions, convolutions with dilation=2 were used: the second, third, and sixth convolutions had dilation=2.

We explore two sets of convolutional layer filters:
\begin{itemize}
    \item \textbf{Slim}: 16, 32, 64, 128, 256, 256, 512, 512
    \item\textbf{Wide}: 32, 64, 128, 256, 512, 512, 1024, 1024
\end{itemize}

After the convolutions, Global Average Pooling was applied, followed by a linear output layer. In these experiments, a dropout after Global Pooling hasn't been used.

All experiments were conducted on an NVIDIA RTX 3090 with identical training parameters.
As part of this research, we aim to find answers to the following questions:
\begin{itemize}

\item  What is the impact of $L_1$ and $L_2$ activation penalties of ConvKAN layers on the model?

\item  What is the impact of $L_1$ regularization of the weights of ConvKAN layers on the model?

\item How does the dropout placement within a ConvKAN layer impact the model? 
Essentially, there are three placements: before the layer (Full), before the calculation of the polynomial basis (Poly), and before applying the weights to the calculated polynomials (Degree).

\item  Since KAN models are supposed to be more robust to noise in the data, can we replace the Dropout layer with additive Gaussian noise as a regularization technique?

\end{itemize}

\begin{table}[!ht]
    \centering
    \resizebox{\textwidth}{!}{\begin{tabular}{l||c|c||c|c}
    \toprule
        \multirow{2}{*}{\textbf{\textbf{Regularization}}} & \multicolumn{2}{c||}{\textbf{Slim}} & \multicolumn{2}{c}{\textbf{Wide}} \\ \cline {2-5}
         & \textbf{Train Accuracy} & \textbf{Val. Accuracy} & \textbf{Train Accuracy} & \textbf{Val. Accuracy} \\ \hline
        None & 73.73 & 61.30 & 86.88 & 67.04 \\ \hline
        $L_1$ Activation, 1e-08 & 74.49 & 61.20 & 86.69 & 66.41 \\ \hline
        $L_1$ Activation, 1e-07 & 74.64 & 61.41 & 86.60 & 66.88 \\ \hline
        $L_1$ Activation, 1e-06 & 74.11 & \textbf{61.74} & 86.73 & 67.09 \\ \midrule \midrule
        $L_2$ Activation, 1e-08 & 73.85 & 61.36 & 86.61 & \textbf{67.49} \\ \hline
        $L_2$ Activation, 1e-07 & 73.96 & 61.55 & 86.78 & 66.81 \\ \hline
        $L_2$ Activation, 1e-06 & 73.10 & 60.29 & 86.99 & 66.30 \\ \midrule \midrule
        $L_1$ Weight, 1e-08 & 74.31 & \textbf{61.39} & 86.41 & 66.79 \\ \hline
        $L_1$ Weight, 1e-07 & 73.48 & 61.08 & 86.41 & 66.79 \\ \hline
        $L_1$ Weight, 1e-06 & 73.60 & 60.96 & 87.10 & 66.83 \\ \midrule \midrule
        Poly Dropout, 0.05 & 68.69 & 60.98 & 84.97 & 66.83 \\ \hline
        Full Dropout, 0.05 & 60.98 & \textbf{63.64} & 79.53 & \textbf{68.93} \\ \hline
        Degree Dropout, 0.05 & 70.12 & 62.91 & 85.01 & 67.75 \\ \hline
        Poly Dropout, 0.15 & 65.43 & 58.63 & 82.07 & 62.26 \\ \hline
        Full Dropout, 0.15 & 43.77 & 56.10 & 63.19 & 66.38 \\ \hline
        Degree Dropout, 0.15 & 64.37 & 62.28 & 81.24 & 68.70 \\ \hline
        Poly Dropout, 0.25 & 61.93 & 51.27 & 80.22 & 59.70 \\ \hline
        Full Dropout, 0.25 & 32.99 & 47.65 & 49.49 & 59.56 \\ \hline
        Degree Dropout, 0.25 & 60.09 & 60.69 & 77.46 & 67.85 \\ \midrule \midrule
        Poly Noise Injection, 0.05 & 70.42 & 62.06 & 84.65 & 67.11 \\ \hline
        Full Noise Injection, 0.05 & 59.75 & \textbf{62.63} & 79.71 & \textbf{69.18} \\ \hline
        Degree Noise Injection, 0.05 & 69.45 & 62.40 & 85.00 & 68.43 \\ \hline
        Poly Noise Injection, 0.15 & 65.75 & 57.64 & 82.78 & 63.82 \\ \hline
        Full Noise Injection, 0.15 & 43.78 & 56.15 & 63.30 & 66.07 \\ \hline
        Degree Noise Injection, 0.15 & 64.52 & 61.78 & 81.33 & 68.23 \\ \hline
        Poly Noise Injection, 0.25 & 63.38 & 53.14 & 80.74 & 59.96 \\ \hline
        Full Noise Injection, 0.25 & 33.54 & 48.24 & 49.58 & 59.70 \\ \hline
        Degree Noise Injection, 0.25 & 60.17 & 60.98 & 77.26 & 67.82 
    \end{tabular}}
    \caption{Regularization study on CIFAR10 dataset}
    \label{reg_results}
\end{table}

Based on the conducted experiments and results presented in Table \ref{reg_results}, the following conclusions can be drawn:
\begin{itemize}
\item  It seems that Full Dropout and Full Noise Injection are the best options for regularization, helping to combat model overfitting.
\item  $L_1$/$L_2$ activation penalties, as well as $L_1$ weight penalty, slightly improve the situation, but not significantly.
\item  In a wider model, the impact of regularization is greater.
\end{itemize}

\subsubsection{Scaling KANs}
\label{sec:scaling}
In classical convolutional networks, we have two major options for scaling up models: we can go deeper and stack more layers, or we can go wider and expand the number of convolutional filters. There are other ways to scale up models, like leveraging a mixture of experts.

KAN Convs with Gram polynomials as basis functions provide us with another possibility for scaling: instead of inflating channel numbers or adding new layers, we could increase the degree of polynomials.

During the experiments, we used the same augmentations as in the previous section, NoiseInjection in Full positions for regularization with p = 0.05 and linear dropout with p = 0.05.

The baseline model was chosen to be simple networks with 8, 12, and 16 convolutional layers with Gram polynomials as basis functions. To reduce the feature's spatial dimensions, convolutions with dilation=2 were used: the second, third, and sixth convolutions had dilation=2 for 8 and 12 layers models, and the second, fourth, and eighth for 16 layers models.

The models have the following sets of convolutional layer filters with a width scale equal to 1:
\begin{itemize}
    \item \textbf{8 layers}: 16, 32, 64, 128, 256, 256, 512, 512
    \item \textbf{12 layers}: 16, 32, 64, 128, 256, 256, 256, 256, 256, 512, 512, 512
    \item \textbf{16 layers}: 16, 16, 32, 32, 64, 64, 128, 128, 256, 256, 256, 256, 512, 512, 512, 512
\end{itemize}

\begin{figure}[!ht]
    \centering
    \includegraphics[width=1.0\linewidth]{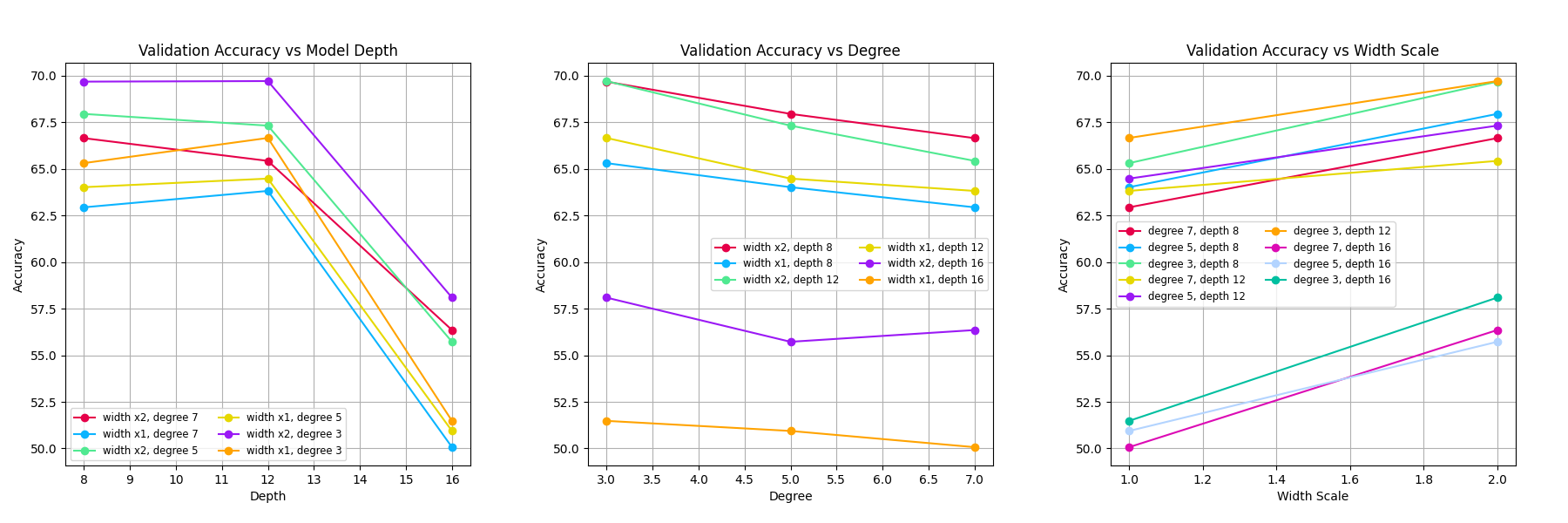}
    \caption{Scaling experiments on CIFAR100 dataset: Accuracy vs. Model depth (left), Accuracy vs. Model degree (center), and Accuracy vs Model width (right).}
    \label{fig:enter-label}
\end{figure}

Our findings indicate that scaling model width yields better performance than increasing model depth. Additionally, there is no observed benefit in scaling the Gram's degree of the model. However, it is noteworthy that the number of trainable parameters increases rapidly in any configuration, whether depth, width, or degree is scaled. 
The lack of benefits from depth and degree scaling may be attributed to the relatively small dataset size. It is plausible that larger datasets could produce different outcomes. 

\subsubsection{Hyperparameters tuning}
\label{sec:param_tuning}

To identify an optimal set of hyperparameters and mitigate the risk of overfitting to the test set, we partitioned the CIFAR100 training dataset into new training and validation sets in an 80/20 ratio. Following the completion of the hyperparameter search, we trained the model on the entire CIFAR100 training set and evaluated it on the complete test set. This study aims to determine effective hyperparameters for eight-layer models. The search space and optimal parameters were established after 50 optimization runs, with the best parameters achieving an accuracy of 61.85

\begin{itemize}
\item  $L_1$ activation penalty, optimal value $10^{-7}$
\item  $L_2$ activation penalty, optimal value $10^{-6}$
\item  $L_1$ weight decay, optimal value $0$
\item  Dropout rate before the output layer, optimal value $0.1456351951990277$
\item  Dropout/Noise Injection, Full placement, optimal value: Full
\item  “Dropout type”: use either Dropout or Noise Injection inside ConvKAGN layers optimal value: Noise Injection
\item  Width scale: parameter that expands the number of channels, optimal value $6$
\item  Degree of Gram polynomials, optimal value $3$
\item  Adam weights decay, optimal value $6.579785489783022 \cdot 10^{-6}$
\item  Learning rate, optimal value $0.000779538356958937$
\item  Learning rate power: parameter that controls the learning rate scheduler, optimal value $1.1275350538654738$
\item  Label smoothing, optimal value $0.1823706816166831$
\end{itemize}

\begin{figure}[!ht]
  \begin{minipage}[b]{.45\linewidth}
    \centering
    \includegraphics[width=0.75\linewidth]{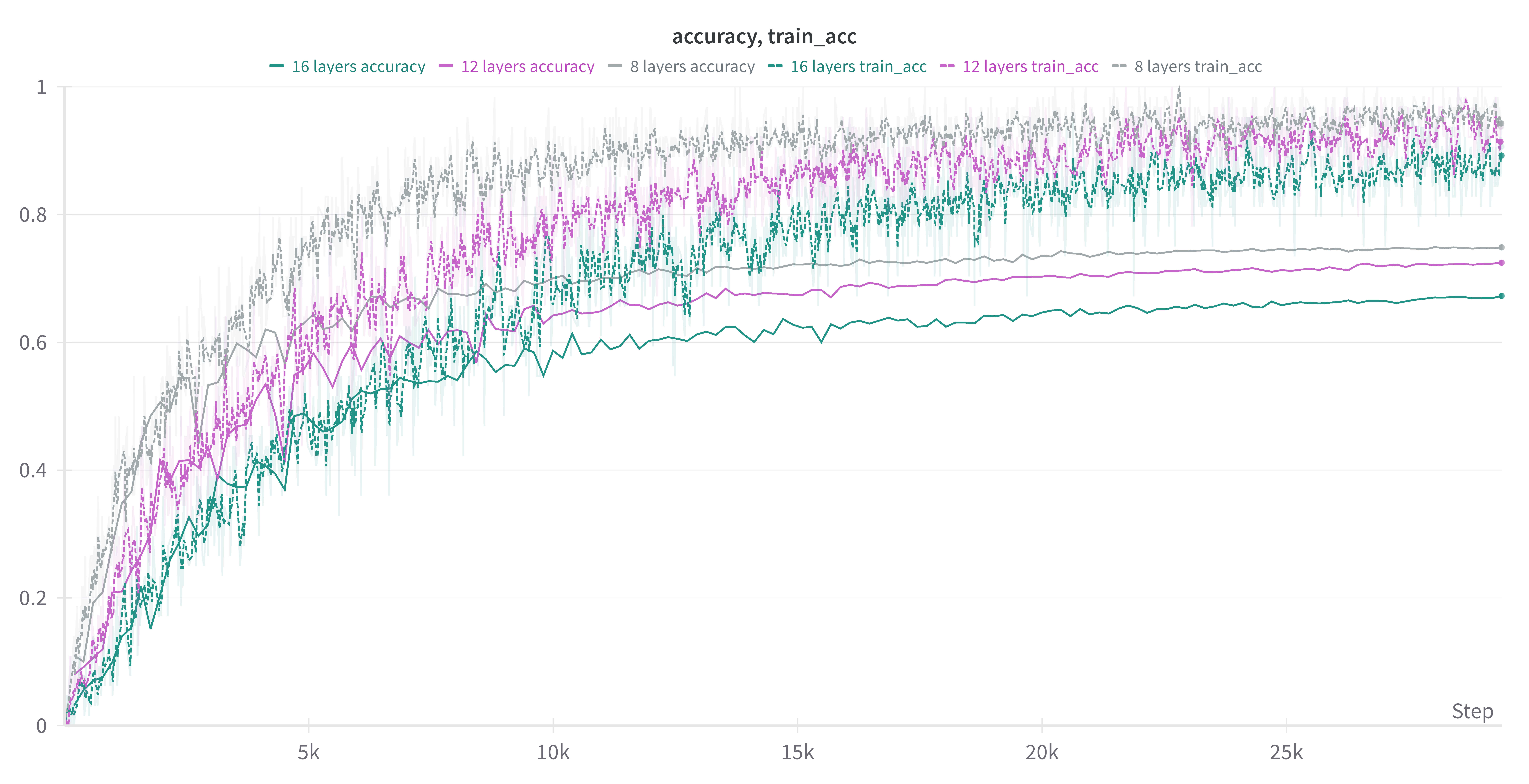}
    \captionof{figure}{Train and validation accuracy curves on CIFAR100, with hyperparameters discovered by hyperparameters otimization}
    \label{fig:hopt_pic}
  \end{minipage}\hfill
  \begin{minipage}[b]{.45\linewidth}
    \centering
    \begin{tabular}{c|c|c}
    \toprule
        \textbf{Val. Accuracy} & \textbf{Model} & \textbf{Parameters} \\ \hline
        \textbf{74.87} & 8 layers & 203.59M \\ \hline
         72.48 & 12 layers & 389.39M \\ \hline
         67.28 & 16 layers & 477.83M \\
    \end{tabular}
    \captionof{table}{Train and validation accuracy on CIFAR100, with hyperparameters discovered by hyperparameters otimization}
    \label{tab:hopt_table}
  \end{minipage}
\end{figure}

The results of the models on this parameter set are presented in Table \ref{tab:hopt_table} and Fig.\ref{fig:hopt_pic}.


The model trained with optimal parameters significantly outperforms the default configuration. Notably, increasing model depth adversely affects performance, which may be due to the extreme model size (nearly half a billion parameters) or issues related to vanishing gradients in very deep networks. However, several established methods, such as ResNet-like and DenseNet-like architectures, can mitigate these issues. The primary challenge remains managing the overwhelming number of parameters in these models. Addressing this challenge is essential for improving scalability and performance.

\subsection{Bottleneck Kolmogorov-Arnold Convolutional on CIFAR100 and Tiny-Imagenet}
\label{sec:cifar100}
\subsubsection{CIFAR100}
In this section, we conduct a series of experiments on the CIFAR-100 dataset using Bottleneck KAGN convolutional layers. The training parameters are based on the hyperparameter optimization results discussed in Section \ref{sec:param_tuning}, all models were trained 200 full epochs. We utilize simple models with 8, 12, and 16 layers, described in Section \ref{sec:scaling}, a tiny DenseNet model as described in \cite{Abai2019DenseNetMF} with all convolutional layers replaced by Bottleneck KAGN convolutions, and ResNet-like models, also with convolutional layers replaced by Bottleneck KAGN convolutional or KAGN convolutional layers. This approach aims to evaluate the performance and scalability of the Bottleneck KAGN layers across different architectures. We also provide results for the Mixture of Experts versions of BottleNeck KAGN convolutional layers, described in Section \ref{sec:bottleneck_kans}, with 2 active experts out of 8.

Tables \ref{tab:bottleneck_kagn_cifar100} and \ref{tab:hopt_table} demonstrate that models using Bottleneck KAGN convolutions perform comparably to those using standard KAGN convolutions but with significantly fewer parameters. However, increasing the number of layers generally degrades model quality. The ResNet18 and ResNet34 variants illustrate that incorporating residual connections mitigates this issue, with ResNet34 outperforming ResNet18 for both KAGN and Bottleneck KAGN convolutions. This indicates that adopting ResNet-like and DenseNet-like approaches can effectively address the challenge of scaling models in depth.

\begin{figure}[!ht]
  \begin{minipage}[b]{.45\linewidth}
    \centering
    \begin{tabular}{l|c}
    \toprule
        \textbf{Model} & \textbf{Accuracy} \\ \hline
        Dspike ResNet-18 \cite{NEURIPS2021_c4ca4238} & 74.24 \\ \hline
        ELU ResNet \cite{10.1145/2983402.2983406} & 73.5 \\ \hline
        OTTT \cite{xiao2022online} & 71.05 \\ \hline
        WaveMix-Lite-256/7 \cite{jeevan2024wavemix} & 70.20 \\ \hline
        IM-Loss (VGG-16) \cite{NEURIPS2022_010c5ba0} & 70.18 \\ \hline
        ResNet18 (modified) \cite{ai4020018} & 66 \\ \midrule \midrule
        ResKAGNet34 (Ours)                & \textbf{0.7814} \\ \hline
        ResKAGNet34 Bottleneck (Ours)      & 0.7711 \\ \hline
        ResKAGNet34 Bottleneck MoE (Ours)  & 0.7166 \\ \hline
        ResKAGNet18 (Ours)                 & 0.7656 \\ \hline
        ResKAGNet18 Bottleneck (Ours)      & 0.7643 \\ \hline
        ResKAGNet18 Bottleneck MoE (Ours)  & 0.7164 \\ \hline
        Tiny DenseKAGNet (Ours)            & 0.7777 \\ \hline
        DenseKAGNet BottleNeck (Ours)      & 0.769  \\ \hline
        DenseKAGNet BottleNeck MoE (Ours)  & 0.7623 \\ \hline
        16 layers, BottleNeck MoE (Ours)   & 0.6953 \\ \hline 
        16 layers, BottleNeck (Ours)       & 0.7065 \\ \hline
        12 layers, BottleNeck MoE (Ours)   & 0.7165 \\ \hline
        12 layers, BottleNeck (Ours)       & 0.7241 \\ \hline
        8 layers, BottleNeck MoE (Ours)    & 0.7439 \\ \hline
        8 layers, BottleNeck (Ours)        & 0.7483 \\ \hline
    \end{tabular}
    \captionof{table}{BottleneckKAGN Convolutional networks, CIFAR 100}
    \label{tab:bottleneck_kagn_cifar100}
  \end{minipage}\hfill
  \begin{minipage}[b]{.45\linewidth}
    \centering
    \begin{tabular}{l|c}
    \toprule
        \textbf{Model} & \textbf{Accuracy} \\ \hline
        Tiny DenseNet \cite{Abai2019DenseNetMF} & 60.0 \\ \hline
        ResNet-18 (AutoMix) \cite{10.1007/978-3-031-20053-3_26} & 67.33 \\ \hline
        ResNeXt-50(AutoMix) \cite{10.1007/978-3-031-20053-3_26}  & \textbf{70.72} \\ \midrule \midrule
        Tiny DenseKAGNet (Ours)     & 66.07  \\ \hline
        Tiny DenseKAGNet, BottleNeck (Ours)     & 65.9  \\ \hline
        Tiny DenseKAGNet, BottleNeck MoE (Ours) & 64.82 \\ \hline
        VGG19-like, MoE  (Ours)                  & 63.72 \\ \hline
        VGG19-like (Ours)                    & 48.59 \\ \hline
        VGG16-like, MoE (Ours)                & 64.27 \\ \hline
        VGG16-like (Ours)                      & 52.55 \\ \hline
        VGG13-like, MoE (Ours)                  & 64.57 \\ \hline
        VGG13-like (Ours)                       & 52.57 \\ \hline
        VGG11-like, MoE (Ours)                  & 63.22 \\ \hline
        VGG11-like (Ours)                       & 51.22 \\ \hline
    \end{tabular}
    \captionof{table}{BottleneckKAGN Convolutional networks, Tiny Imagenet}
    \label{tab:bottleneck_kagn_tiny_imagenet}
  \end{minipage}
\end{figure}

\subsubsection{Tiny ImageNet}
In this section, we conduct a series of experiments on the Tiny ImageNet dataset. The training parameters are based on the hyperparameter optimization results discussed in \ref{sec:param_tuning}, all models were trained 200 full epochs. We employ VGG-like models where hidden linear layers and the final MaxPool layer are replaced with a Global Average Pooling and a single output layer. Additionally, we use a tiny DenseNet model, as described in \cite{Abai2019DenseNetMF}, with all convolutional layers replaced by Bottleneck KAGN convolutionals. We also perform experiments with Mixture of Experts (MoE) models featuring two active experts out of a total of eight.

The results, presented in Table \ref{tab:bottleneck_kagn_tiny_imagenet}, indicate that VGG-like models decrease in accuracy as the number of layers increases. In contrast, the MoE version shows significantly better performance on this dataset, suggesting that MoE layers can be an effective approach for scaling KAGN-based models in width.

Also, proposed models, VGG-like MoE and Tiny DenseNet, with Gram polynomials KAN convolutions outperform Tiny DenseNet \cite{Abai2019DenseNetMF} by a significant margin on the Tiny Imagenet dataset.

\subsection{Self-KAGNtention}
\label{sec:selfatt}

\begin{figure}[!ht]
    \centering
    \includegraphics[width=0.75\linewidth]{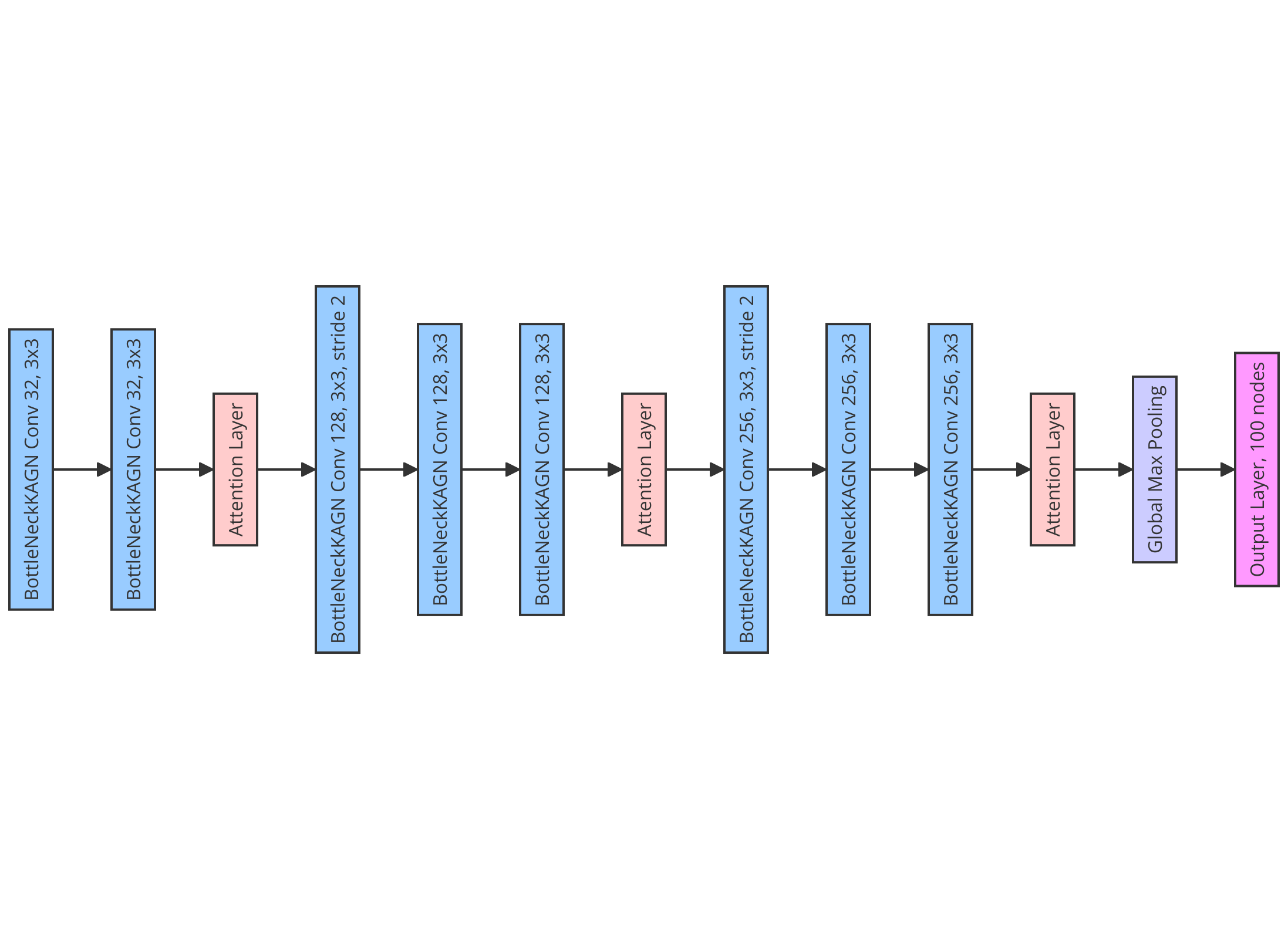}
    \caption{Model architecture with attention blocks. As an attention, one could use either SelfKAGNtention or BottleNeckKAGN Focal Modulation layer.}
    \label{fig:model-attention}
\end{figure}

\begin{figure}[!ht]
    \centering
    \includegraphics[width=0.75\linewidth]{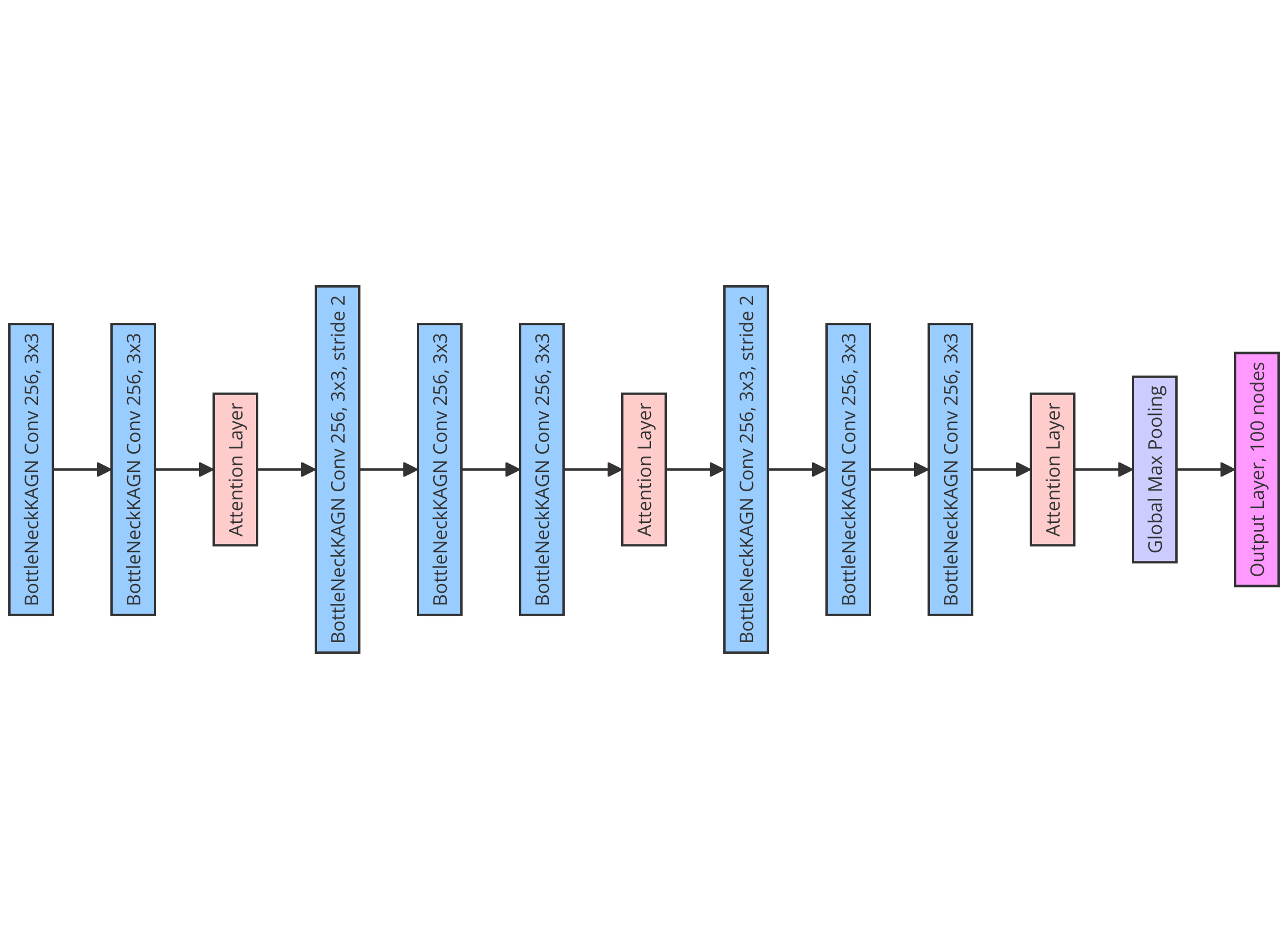}
    \caption{Fixed model architecture with attention blocks. As an attention, one could use either SelfKAGNtention or BottleNeckKAGN Focal Modulation layer. The main difference between this model and Fig.\ref{fig:model-attention} is the number of filters in each layer. The fixed version uses the same filter numbers across all layers.}
    \label{fig:model-attention-fixed}
\end{figure}

In this section, we empirically investigate the performance of BottleNeck SelfKAGNtention layers and BottleNeckKAGN Focal Modulation. We base our experiments on an architecture with eight convolutional layers, supplemented with three Self-Attention or Focal Modulation layers. The network (see Fig.\ref{fig:model-attention} and \ref{fig:model-attention-fixed}) begins with two BottleNeckKAGN convolution layers, each with 32 filters (256 filters in the Fixed version), both using a 3x3 kernel. This is followed by an attention layer. The next section consists of three BottleNeckKAGN convolution layers, each with 128 filters (256 filters in the Fixed version) and a 3x3 kernel, with the first layer using a stride of 2. Another attention layer is added after these layers. The subsequent section includes three BottleNeckKAGN convolution layers with 256 filters and a 3x3 kernel, with the first layer using a stride of 2. A third attention layer follows. The network concludes with a global max pooling layer and an output layer with 100 nodes.

We use the training and regularization parameters found in Section \ref{sec:param_tuning}, categorical cross-entropy as the loss function, and the AdamW optimizer \cite{Loshchilov2017DecoupledWD}. The models were trained for 150 epochs with a batch size of 256. For Self-Attention layers, we employ bottleneck convolutions with 16 and 64 filters in the first two attention layers to reduce memory and computational resource usage.

We explore two model versions: one with an increasing number of filters with depth and another with a constant number of filters, inspired by transformer architectures. The experimental results on CIFAR-100, presented in Table \ref{tab:attention}, and compared with Table \ref{tab:hopt_table}, show that the combination of attention mechanisms with bottleneck convolutions outperforms models without attention while maintaining a significantly lower number of parameters.

\begin{table}[!ht]
\centering
\begin{tabular}{|l|c|c|c|}
\toprule
\textbf{Model}                                                   & \textbf{Width scale} & \textbf{Accuracy, CIFAR-100} & \textbf{Parameters, M} \\ \midrule
SelfKAGNtention                                                  & 1                    & 71.4                         & 4.9                    \\ \hline
Fixed SelfKAGNtention                                            & 1                    & \textbf{77.82}                        & 8.7                    \\ \hline
FocalKAGNtention                                                 & 1                    & 71.24                        & 4.3                    \\ \hline
Fixed FocalKAGNtention                                           & 1                    &                  75.99            &     
                7.4   \\ \hline
SelfKAGNtention                                                  & 3                    & 77.69                        & 43.7                   \\ \hline
Fixed SelfKAGNtention & 3                    & 69.72                        & 78.3                   \\ \hline
FocalKAGNtention                                                 & 3                    &                 76.46             &   37.8                     \\ \hline
Fixed FocalKAGNtention                                           & 3                    &                  7467            &       63.2            \\
\bottomrule
\end{tabular}
\captionof{table}{BottleneckKAGN Convolutional with Attention layers, CIFAR-100 dataset}
\label{tab:attention}
\end{table}

\subsection{Imagenet1k}
\label{sec:in1k}
In this section, we provide results on ImageNet1k \cite{journals/corr/RussakovskyDSKSMHKKBBF14} dataset. We have tested several VGG-like \cite{simonyan2015a} models with several modifications. We replaced the last MaxPolling layers and two hidden fully connected layers by GlobalAveragePooling and one output fully connected layer. We also added two extra convolutional layers at the end of the encoder.

\begin{figure}[!ht]
    \centering
    \includegraphics[width=1.0\linewidth]{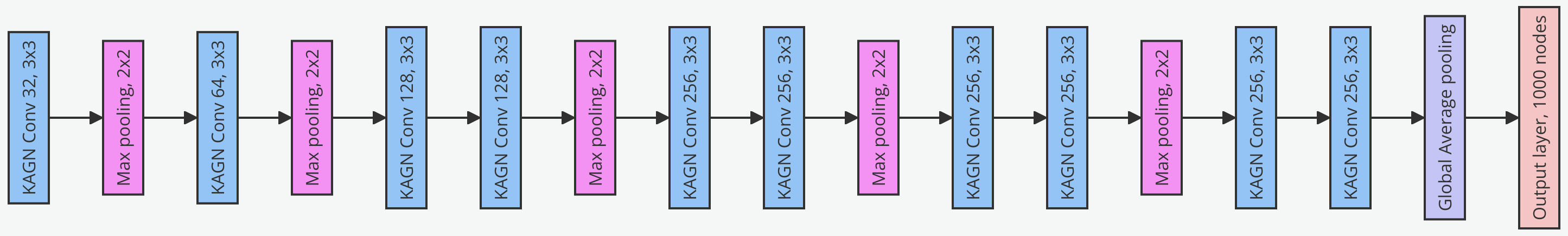}
    \caption{The scheme of the VGG11-like v2 model}
    \label{fig:vgg11-v2}
\end{figure}

\begin{figure}[!ht]
    \centering
    \includegraphics[width=1.0\linewidth]{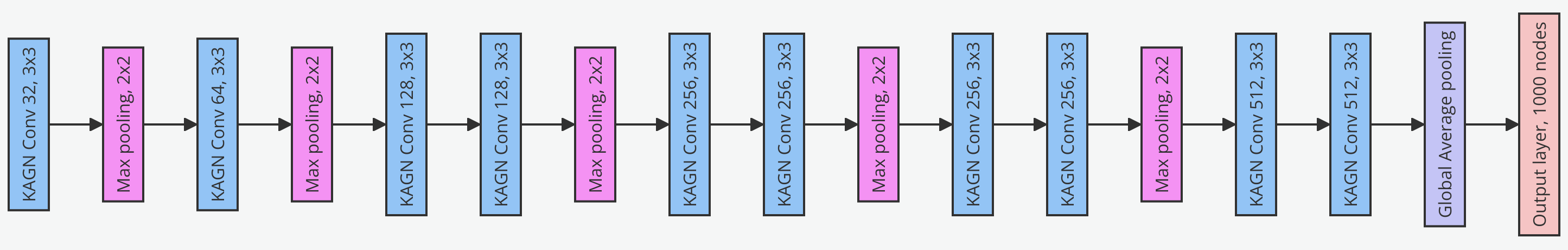}
    \caption{The scheme of the VGG11-like v4 model. Please note the difference between V2 and V4 models. V4 models have wider 2 last convolutional layers.}
    \label{fig:vgg11-v4}
\end{figure}

The model consists of consecutive 10 Gram ConvKAN Layers or Bottleneck Gram ConvKAN Layers with BatchNorm, polynomial degree equals 5, GlobalAveragePooling, and Linear classification head (see Fig.\ref{fig:vgg11-v2} and \ref{fig:vgg11-v4}). The network design starts with a KAGN convolution layer with 32 filters (3x3), followed by a max pooling layer (2x2). This pattern is repeated with a KAGN convolution layer with 64 filters (3x3) and another max pooling layer (2x2). The network then includes two consecutive KAGN convolution layers with 128 filters (3x3), followed by another max pooling layer (2x2). Next, there are two KAGN convolution layers with 256 filters (3x3), another max pooling layer (2x2), and two more KAGN convolution layers with 256 filters for the V2 version or 512 filters for the V4 version (3x3). The network concludes with a global average pooling layer and a dense output layer with 1000 nodes.


We also have tested a model with a Self KAN-attention layer, described in \ref{sec:selfkantention} placed before Global Average pooling.


All models, except one, were trained during 200 full epochs with AdamW optimizer, with learning rate equals 0.0009, $\beta_1=0.9, \beta_2=0.999$, weight decay equals $5 \cdot 10^{-6}$, with 7500 warmup steps and polynomial learning rate scheduler with power $0.3$ and learning rate at the end $10^{-7}$, and batch size equals 32. The model marked with "Opt params" tag was trained with the parameters discovered in Section \ref{sec:param_tuning}, and a polynomial degree equals 3 for this model. The models marked with "BottleNeck" use BottleNeck convolutional layers, described in Section\ref{sec:bottleneck_kans}.

The results of experiments are presented in Table \ref{tab:imagenet1k}. Please, note that the metrics of our experiments are computed on the validation set of ImageNet1k. Metrics on the test set will be added later. According to the acquired results, the VGG11-like model with bottleneck KAGN convolutions outperforms all models from the VGG family, ResNet-18, ResNet-34, and DenseNet121 models. This fact shows the great potential of the KAN convolutions model.

\begin{table}[!ht]
    \centering
    \begin{tabular}{l|c|c}
    \toprule
        \textbf{Model} & \textbf{Accuracy, Top-1} & \textbf{Accuracy, Top-5} \\ \hline
        VGG11 \cite{simonyan2015a} & 69.02 & 88.628 \\ \hline
        VGG11, w BatchNorm \cite{simonyan2015a} & 70.37 & 89.81 \\ \hline
        VGG13 \cite{simonyan2015a} & 69.928 & 89.246 \\ \hline
        VGG13, w BatchNorm \cite{simonyan2015a} & 71.586 & 	90.374 \\ \hline
        VGG16 \cite{simonyan2015a} & 71.592 & 90.382 \\ \hline
        VGG16, w BatchNorm \cite{simonyan2015a} & 73.36 & 	91.516 \\ \hline
        VGG19 \cite{simonyan2015a} & 72.376 & 90.876 \\ \hline
        VGG19, w BatchNorm \cite{simonyan2015a} & 74.218 & 	91.842 \\ \hline
        ResNet18 \cite{7780459} & 69.758 & 89.078 \\ \hline
        ResNet34 \cite{7780459} & 73.314 & 91.42 \\ \hline
        ResNet50 \cite{7780459} & 76.13 & 92.862 \\ \hline
        DenseNet121 \cite{8099726} & 74.434 & 91.972 \\ \hline
        DenseNet161 \cite{8099726} & \textit{77.138} & \textit{93.56} \\ \midrule \midrule
        VGG11-like, v2 (Ours)\tablefootnote{The weights are available on \href{https://huggingface.co/brivangl/vgg_kagn11_v2}{HugginFace}}  & 59.1 &	82.29 \\ \hline
        VGG11-like, v4 (Ours)\tablefootnote{The weights are available on \href{https://huggingface.co/brivangl/vgg_kagn11_v4}{HugginFace}}  & 61.17 & 83.26 \\ \hline
        VGG11-like, v4, BottleNeck (Ours)\tablefootnote{The weights are available on \href{https://huggingface.co/brivangl/vgg_kagn_bn11_v4}{HugginFace}}  & 68.50 & 88.46 \\ \hline
        VGG11-like, v4, BottleNeck, SA (Ours)\tablefootnote{The weights are available on \href{https://huggingface.co/brivangl/vgg_kagn_bn11sa_v4}{HugginFace}} &70.684 & 89.462 \\ \hline
        VGG11-like, v4, BottleNeck, Opt params (Ours)\tablefootnote{The weights are available on \href{https://huggingface.co/brivangl/vgg_kagn_bn11_v4_opt}{HugginFace}}  & \textbf{74.586} & \textbf{92.13} \\ \hline

    \end{tabular}
    \caption{Results on the ImageNet 1K dataset}
    \label{tab:imagenet1k}
\end{table} 


\subsection{Parameter Efficient Finetuning}
\label{sec:peft}

In this section, we use the HAM10000 ("Human Against Machine with 10000 training images") dataset \cite{DVN/DBW86T_2018} to explore the proposed PEFT method for Gram KAN convolution models. The dataset consists of 10015 dermatoscopic images. Cases include a representative collection of all important diagnostic categories in the realm of pigmented lesions: Actinic keratoses and intraepithelial carcinoma / Bowen's disease, basal cell carcinoma, benign keratosis-like lesions, dermatofibroma, melanoma, melanocytic nevi, and vascular lesions. We use the train, validation, and test split hosted on HuggingFace: \href{https://huggingface.co/datasets/marmal88/skin_cancer}{https://huggingface.co/datasets/marmal88/skin\_cancer}. 

We selected this dataset to evaluate the PEFT method due to its requirement for model adaptation to a novel domain not encountered during pretraining on ImageNet1k. This characteristic makes it an exemplary dataset for our experiments, providing a scenario closely aligned with real-world use cases.

We evaluate several setups with the same training procedure and parameters (see Table \ref{tab:peft_res}). We use the VGG11-like, v2 model pretrained on ImageNet1K. All models were trained during 20 full epochs with AdamW optimizer, with learning rate equals 0.0001, $\beta_1=0.9, \beta_2=0.999$, weight decay equals $5 \cdot 10^{-6}$, with 500 warmup steps and polynomial learning rate scheduler with power $0.3$ and learning rate at the end $10^{-6}$, and batch size equals 12.  Because of class imbalance in the dataset, we use Focal Loss \cite{8417976} instead of categorical cross-entropy.

Setups with "Random weights initialization" value in the "Pretrained degree" column mark setups with random weights initialization with the same parameters as the VGG11-like, v2 model. The column "Trainable Degree" describes a set of trainable parameters of $\Tilde{w}_g = \{w_i\}_{i=R}^{N+1}, R > 1$, 6 means that extra polynomial was added to the basis. "Full training" implies that all weights of the encoder were updated during the training process. The "Trainable activation residual" column indicates whether or not $w_b$ weights of residual activation were trainable or not.

As we can see from Table \ref{tab:peft_res}, PEFT setup with trainable 4-th, 5-th and extra degree polynomial-related weights outperforms all other setups, showing the potential of the proposed method, including full weight fine-tuning. Also we should mention that fine-tuning activation residual weights in general leads to worse performance, so it's safe to assume that during PEFT procedure these weights could remain frozen. 

\begin{table}[ht]
\resizebox{\textwidth}{!}{\begin{tabular}{c|c|c|c|c|c|c|c|c|c}
\toprule
\multicolumn{3}{c|}{\textbf{Training setup}}                                                        & \multicolumn{7}{c}{\textbf{Metrics}}                                                                                                                                                                          \\ \hline
\textbf{Pretrained degree}             & \textbf{Trainable Degree} & \textbf{Finetune activation residual} & \textbf{Accuracy} & \textbf{AUC (OvR)} & \textbf{AUC (OvO)} & \textbf{Recall (macro)} & \textbf{Recall (micro)} & \textbf{$F_1$ score (macro)} & \textbf{$F_1$ score (micro)} \\ \midrule \midrule
5                             & 4, 5, 6          & FALSE                        & \textbf{83.89}    & 0.975    & 0.984    & \textbf{0.842}         & \textbf{0.839}         & 0.752            & \textbf{0.839}            \\ \hline
5                             & 4                & FALSE                        & 82.72    & \textbf{0.977}    & 0.983    & 0.838         & 0.827         & \textbf{0.777}            & 0.827            \\ \hline
5                             & 4, 5, 6          & TRUE                         & 81.79    & 0.971    & 0.983    & 0.827         & 0.818         & 0.714            & 0.818            \\ \hline
5                             & 4, 5             & FALSE                        & 80.7     & \textbf{0.977}    & \textbf{0.985}    & 0.827         & 0.807         & 0.75             & 0.807            \\ \hline
5                             & 5                & FALSE                        & 80.16    & 0.974    & 0.98     & 0.802         & 0.802         & 0.758            & 0.802            \\ \hline
5                             & 4, 5             & TRUE                         & 77.28    & 0.966    & 0.974    & 0.769         & 0.773         & 0.68             & 0.773            \\ \hline
5                             & Full Training    & TRUE                         & 76.89    & 0.976    & 0.985    & 0.842         & 0.769         & 0.729            & 0.769            \\ \hline
5                             & 5                & TRUE                         & 75.8     & 0.963    & 0.972    & 0.783         & 0.758         & 0.696            & 0.758            \\ \hline
5                             & 6                & FALSE                        & 74.32    & 0.941    & 0.94     & 0.634         & 0.743         & 0.636            & 0.743            \\ \hline
5                             & 3                & FALSE                        & 74.16    & 0.945    & 0.939    & 0.573         & 0.742         & 0.601            & 0.742            \\ \hline
5                             & 4                & TRUE                         & 73.07    & 0.967    & 0.981    & 0.84          & 0.731         & 0.69             & 0.731            \\ \hline
Random weights initialization & Full Training    & TRUE                         & 71.67    & 0.958    & 0.976    & 0.837         & 0.717         & 0.67             & 0.717            \\ \hline
5                             & 2                & FALSE                        & 71.28    & 0.927    & 0.924    & 0.507         & 0.713         & 0.529            & 0.713            \\ \hline
5                             & 6                & TRUE                         & 69.18    & 0.938    & 0.947    & 0.69          & 0.692         & 0.639            & 0.692            \\ \hline
5                             & 5, 6             & FALSE                        & 67.78    & 0.936    & 0.938    & 0.55          & 0.678         & 0.54             & 0.678            \\ \hline
5                             & 2                & TRUE                         & 66.69    & 0.928    & 0.939    & 0.634         & 0.667         & 0.599            & 0.667            \\ \hline
5                             & 5, 6             & TRUE                         & 66.69    & 0.939    & 0.946    & 0.577         & 0.667         & 0.564            & 0.667            \\ \hline
5                             & 3                & TRUE                         & 58.13    & 0.944    & 0.955    & 0.678         & 0.581         & 0.622            & 0.581            \\ \hline
5                             & Fixed encoder    & FALSE                        & 48.25    & 0.864    & 0.87     & 0.53          & 0.482         & 0.361            & 0.482   \\
\bottomrule

\end{tabular}}
\caption{PEFT results on HAM10000 dataset}
\label{tab:peft_res}
\end{table}

\subsection{Segmentation}
\label{sec:seg}
In this section, we provide empirical evaluation for U-Net-like segmentation models based on KAGN Convolutional layers on BUSI, GlaS, and CVC-ClinicDB datasets.

The BUSI dataset \cite{al2020dataset} consists of ultrasound images depicting normal, benign, and malignant breast cancer cases, along with their corresponding segmentation maps. In our study, we utilized 647 ultrasound images representing both benign and malignant breast tumors, all consistently resized to 256 × 256 pixels. This dataset provides a comprehensive collection of images that assist in detecting and differentiating various types of breast tumors, offering valuable insights for medical professionals and researchers.

The GlaS dataset \cite{valanarasu2021medical} includes 612 Standard Definition (SD) frames from 31 sequences, each with a resolution of 384 × 288 pixels, collected from 23 patients. This dataset is associated with the Hospital Clinic in Barcelona, Spain. The sequences were recorded using Olympus Q160AL and Q165L devices, paired with an Extra II video processor. For our study, we specifically used 165 images from the GlaS dataset, all resized to 512 × 512 pixels.

The CVC-ClinicDB dataset \cite{valanarasu2021medical}, also known as "CVC," is a publicly accessible resource for polyp diagnosis within colonoscopy videos. It comprises 612 images, each with a resolution of 384 × 288 pixels, extracted from 31 distinct colonoscopy sequences. These frames offer a diverse array of polyp instances, making them particularly useful for developing and evaluating polyp detection algorithms. To ensure consistency across different datasets in our study, all images from the CVC-ClinicDB dataset were uniformly resized to 256 × 256 pixels.

We explore 3 different U-net-like models: first is a U-net with convolutions replaced by KAGN Convolutional layers, $U^2$-net-like model \cite{QIN2020107404}, again with convolutions replaced by KAGN Convolutional layers, and $U^2$-net small model, were all filter number were the same across all hidden layers and equal 16 multiplied by the width scale parameter.

For the BUSI, GlaS, and CVC datasets, the batch size was set to 4 and the learning rate was 1e-4, all other parameters were the same as discussed in Section \ref{sec:param_tuning}. The loss function was chosen to be a combination of binary cross entropy and dice loss. We randomly split each dataset into 80\% training and 20\% validation subsets.

\begin{figure}[!ht]
    \centering
\resizebox{\textwidth}{!}{    \includegraphics[width=0.75\linewidth]{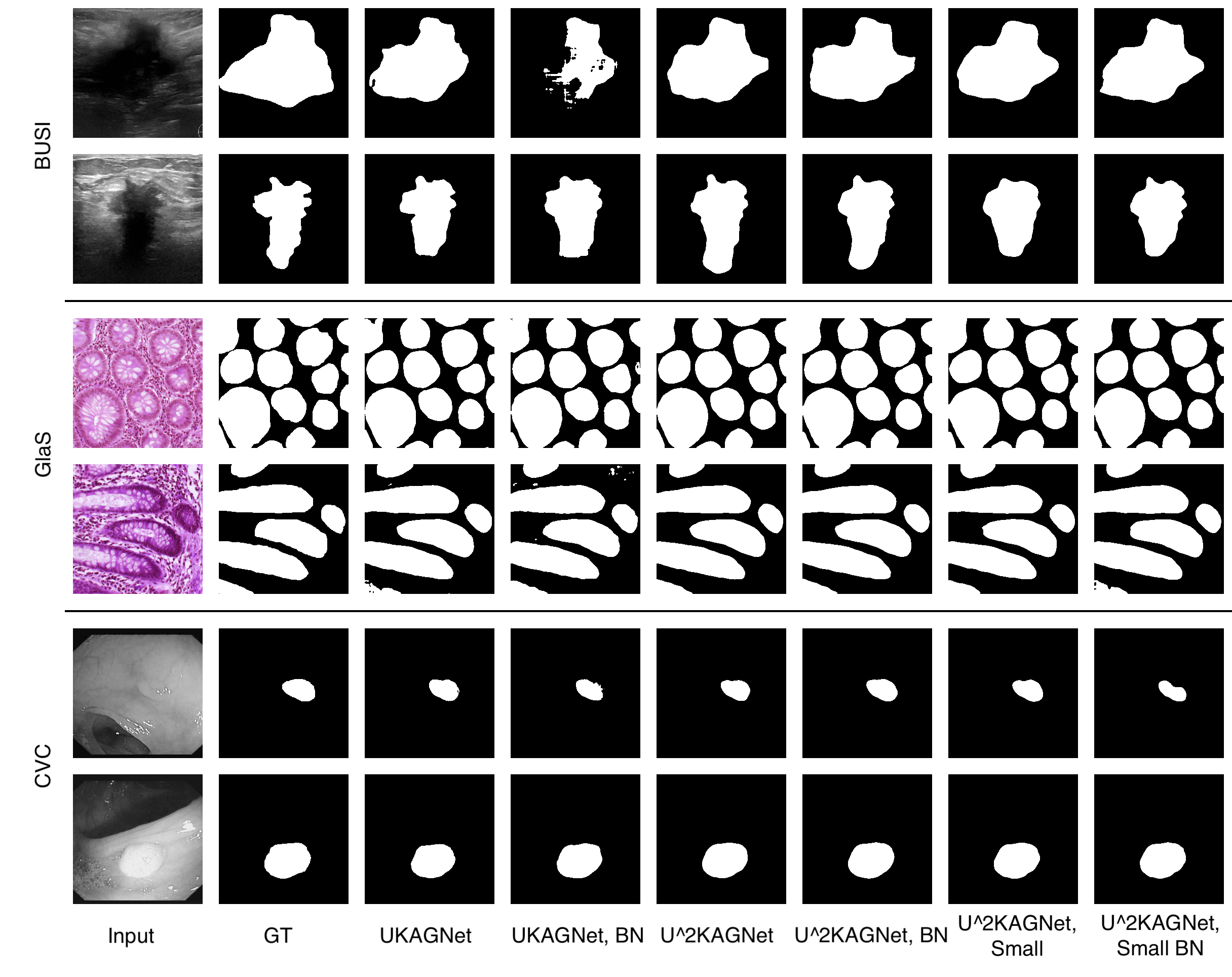}}
    \caption{Examples of model predictions on three diverse medical image datasets}
    \label{fig:seg}
\end{figure}

The results are presented in Table \ref{tab:exp_seg}. Our approach outperforms all state-of-the-art methods, including the novel U-KAN model \cite{li2024ukanmakesstrongbackbone}. Qualitative results of the models are provided in Fig.\ref{fig:seg}.

\begin{table}[htbp]
  \centering
  \makeatletter\def\@captype{table}\makeatother
  
  \resizebox{\linewidth}{!}{
   \setlength{\tabcolsep}{1.8mm}
    \begin{tabular}{lcccccc}
    \toprule
    \multirow{2}[4]{*}{Methods} & \multicolumn{2}{c}{BUSI~\cite{al2020dataset}}    & \multicolumn{2}{c}{GlaS~\cite{valanarasu2021medical}}    & \multicolumn{2}{c}{CVC~\cite{bernal2015wm}} \\
\cmidrule{2-7}                 & IoU↑         & F1↑          & IoU↑         & F1↑          & IoU↑         & F1↑ \\
    \midrule
    U-Net~\cite{ronneberger2015unet}      &  57.22 & 71.91 & 86.66 & 92.79  & 83.79  & 91.06 \\
    Att-Unet~\cite{oktay2018attention}     & 55.18   & 70.22   & 86.84   & 92.89    & 84.52  & 91.46 
 \\
    U-Net++~\cite{zhou2018unet++}       & 57.41    & 72.11    & 87.07   & 92.96   & 84.61 
  & 91.53 
 \\
    U-NeXt~\cite{valanarasu2022unext}        & 59.06    & 73.08    & 84.51    & 91.55  &74.83 & 85.36 
 \\
    Rolling-UNet~\cite{liu2024rolling}    & 61.00   & 74.67    & 86.42    & 92.63   & 82.87   & 90.48 
 \\
    U-Mamba~\cite{ma2024u}      &  61.81 & 75.55  & 87.01 & 93.0 & 84.79  & 91.63  \\

   U-KAN \cite{li2024ukanmakesstrongbackbone} &     63.38  & 76.40  & 87.64  & 93.37 & 85.05 & 91.88       \\ \midrule
   $U^2$-KAGNet, Small (Ours)            & 59.44     & 74.56           & \textbf{89.13}     & \textbf{94.25}           & 85.62    & 92.26          \\
    $U^2$-KAGNet, BottleNeck, Small (Ours) & 62.95     & 77.26           & 85.99     & 92.47           & 87.61    & 93.39          \\
    $U^2$-KAGNet, BottleNeck (Ours)        & 55.91     & 71.72           & 87.73     & 93.47           & \textbf{88.86}    & \textbf{94.1}           \\
    $U^2$-KAGNet (Ours)                    & 58.13     & 73.52           & 88.14     & 93.7            & 87.07    & 93.09          \\
    UKAGNet (Ours)                      & \textbf{63.45}     & \textbf{77.64}           & 87.31     & 93.23           & 76.85    & 86.91    \\      

    \bottomrule
    \end{tabular}
    }
  \caption{
Comparison with state-of-the-art segmentation models on three heterogeneous medical datasets.
  }
  \label{tab:exp_seg}
\end{table}

\section{Ablation Study}

In this section, we present the results of an ablation study of the Bottleneck Kolmogorov-Arnold convolutional layers with Gram polinomials. The baseline sequence for convolutional KAN models includes an activation residual summed with a nonlinearity $\Tilde{\varphi}$, followed by a normalization layer (e.g., batch norm, instance norm), and a SiLU activation. The ablation experiments involve excluding one or more elements from this sequence: activation residual, normalization layer, nonlinearity, and replacing the linear bottleneck with a KAN-based bottleneck layer. Experiments were conducted on the MNIST \cite{deng2012mnist}, CIFAR-10, CIFAR-100, and Fashion MNIST \cite{xiao2017/online} datasets, and results are shown in Table \ref{tab:ablation}.

The results show that except for MNIST, the KAN-based bottleneck leads to training collapse or significant accuracy degradation, and using activation residual degrades performance in half of the cases. This suggests that activation residual may be redundant in some cases and warrants further investigation. Nevertheless, we propose retaining the convolutional layer scheme with activation residual, as the bottleneck approach can lead to information loss, and activation residual can help in its recovery.

\begin{table}[ht]
\resizebox{\linewidth}{!}{\begin{tabular}{l|l|c|c|c|c}
\toprule
Model & Setup                                      & \makecell{Accuracy, \\ MNIST} & \makecell{Accuracy, \\ Fashion-MNIST} & \makecell{Accuracy, \\ CIFAR-10} & \makecell{Accuracy, \\ CIFAR-100} \\ \midrule
8 layers & activation skip, linear bottleneck         & 0.1135          & 0.8948                  & 0.7392             & 0.01                \\
8 layers & activation skip, linear bottleneck, activation       & 0.1135          & 0.1                     & 0.1                & 0.01                \\
8 layers & activation skip, KAN bottleneck, batch norm, activation    & 0.1135          & 0.1                     & 0.1                & 0.01                \\
8 layers & linear bottleneck, batch norm, activation           & 0.9967          & \textbf{0.9424}         & \textbf{0.8999}    & \textbf{0.6672}     \\
8 layers & activation skip, linear bottleneck, batch norm, activation & \textbf{0.9969} & 0.9348                  & 0.8923             & 0.6489              \\ \midrule
4 layers & activation skip, linear bottleneck            & 0.9552          & 0.815                   & 0.5438             & 0.2614              \\
4 layers & activation skip, linear bottleneck, activation       & 0.9858          & 0.83                    & 0.626              & 0.2801              \\
4 layers & activation skip, KAN bottleneck, batch norm, activation    & \textbf{0.9949} & 0.8879                  & 0.7447             & 0.4417              \\
4 layers & linear bottleneck, batch norm, activation          & 0.9935          & \textbf{0.9108}         & \textbf{0.8021}    & 0.4921              \\
4 layers & activation skip, linear bottleneck, batch norm, activation & 0.9941          & 0.9016                  & 0.7962             & \textbf{0.4951}    \\
\bottomrule
\end{tabular}}
\caption{
Ablation study results
  }
\label{tab:ablation}
\end{table}

\section{Design Principles}

Summarizing the experiments conducted, we propose the following preliminary design principles for Kolmogorov-Arnold convolutional networks. It is important to note that our experiments were not exhaustive, and these principles may be revised with new data.

\begin{itemize}
\item We recommend using Gram polynomials for $\Tilde{\varphi}(x)$. Our experiments indicate this choice excels in both quality metrics and the number of trainable parameters.
\item For scaling Kolmogorov-Arnold convolution-based models, we suggest using the bottleneck version of the layers, which significantly reduces the number of trainable parameters without substantial loss in performance compared to the non-bottleneck version.
\item Increasing model width generally performs better than increasing depth, as shown by our experiments with simple sequential models. The Mixture of Experts versions of the bottleneck convolution effectively scales model width without a significant increase in inference and training costs.
\item Preliminary findings suggest that DenseNet-like architectures could serve as a strong foundation for constructing very deep Kolmogorov-Arnold convolutional networks.
\item Our experiments demonstrate that Self KAGNtention layers can enhance the performance of Kolmogorov-Arnold convolutional models.
\item Kolmogorov-Arnold convolutions perform exceptionally well in segmentation tasks, with U2Net recommended as a starting architecture for further research.
\item As shown in Section \ref{sec:reg}, $L_1$ and $L_2$ activation regularization, and Noise Injection layers before the Kolmogorov-Arnold convolutional layer are effective regularization techniques.
\end{itemize}

\section{Conclusion}
This paper explores the integration of Kolmogorov-Arnold Networks (KANs) into convolutional neural network architectures, presenting novel approaches and modifications to enhance their performance and efficiency in computer vision tasks. Our work introduces Bottleneck Convolutional Kolmogorov-Arnold layers, a parameter-efficient design that reduces memory requirements and mitigates overfitting issues. Additionally, we propose a parameter-efficient fine-tuning algorithm that significantly decreases the number of trainable parameters needed for adapting pre-trained models to new tasks.

Through extensive empirical evaluations on various datasets, including MNIST, CIFAR10, CIFAR100, Tiny ImageNet, ImageNet1k, HAM10000, BUSI, GlaS, and CVC-ClinicDB, we demonstrate that KAN-based convolutional models can achieve state-of-the-art results in both classification and segmentation tasks. Our experiments highlight the effectiveness of Gram polynomials as the basis function for KANs, the advantages of scaling model width over depth, and the potential of DenseNet-like architectures for very deep networks.

We further show that incorporating Self KAGNtention layers enhances model performance, particularly in complex tasks, and provide design principles for constructing successful KAN convolutional models. Our proposed models not only outperform traditional convolutional networks but also offer a promising direction for future research in optimizing neural network architectures for computer vision applications.

Overall, our findings emphasize the potential of Kolmogorov-Arnold Networks in advancing the capabilities of convolutional neural networks, paving the way for more efficient and effective deep learning models. Future work will focus on refining these approaches and exploring their applications in other domains, as well as investigating additional regularization techniques and optimization strategies to further enhance model performance.

\bibliographystyle{unsrt}

\bibliography{references}  

\end{document}